\documentclass[runningheads]{llncs}
\usepackage{pdfpages}
\usepackage{graphicx}
\graphicspath{ {images/} }
\usepackage{dsfont}
\usepackage{bm}
\usepackage[utf8]{inputenc} 
\usepackage[T1]{fontenc}    
\usepackage{hyperref}       
\usepackage{url}            
\usepackage{booktabs}       
\usepackage{amsfonts}       
\usepackage{nicefrac}       
\usepackage{microtype}      
\usepackage{amssymb}
\usepackage{amsmath}
\usepackage{multirow}
\usepackage[]{algorithm2e}
\usepackage{afterpage}
\usepackage{float}
\usepackage{listings}

\begin{document}

\title{Fake News Detection by means of Uncertainty Weighted Causal Graphs}

%

\author{Eduardo C. Garrido-Merch\'an\inst{1} \and
Cristina Puente\inst{2} \and
Rafael Palacios\inst{2}}

\institute{Universidad Aut\'onoma de Madrid, Francisco Tom\'as y Valiente 11, Madrid, Spain
\email{eduardo.garrido@uam.es} \and
Escuela T\'ecnica Superior de Ingenier\'ia ICAI, Universidad Pontificia Comillas, Madrid, Spain
\email{cristina.puente@icai.comillas.edu} \and
Escuela T\'ecnica Superior de Ingenier\'ia ICAI, Universidad Pontificia Comillas, Madrid, Spain
\email{palacios@comillas.edu}}

\maketitle
\begin{abstract}%
Society is experimenting changes in information consumption, as new information channels such as social networks let people share news that do not necessarily be trust worthy. Sometimes, these sources of information produce fake news deliberately with doubtful purposes and the consumers of that information share it to other users thinking that the information is accurate. This transmission of information represents an issue in our society, as can influence negatively the opinion of people about certain figures, groups or ideas. Hence, it is desirable to design a system that is able to detect and classify information as fake and categorize a source of information as trust worthy or not. Current systems experiment difficulties performing this task, as it is complicated to design an automatic procedure that can classify this information independent on the context. In this work, we propose a mechanism to detect fake news through a classifier based on weighted causal graphs. These graphs are specific hybrid models that are built through causal relations retrieved from texts and consider the uncertainty of causal relations. We take advantage of this representation to use the probability distributions of this graph and built a fake news classifier based on the entropy and KL divergence of learned and new information. We believe that the problem of fake news is accurately tackled by this model due to its hybrid nature between a symbolic and quantitative methodology. We describe the methodology of this classifier and add empirical evidence of the usefulness of our proposed approach in the form of synthetic experiments and a real experiment involving lung cancer. 
\end{abstract}
\section{Introduction}
There is a trend in our society to read more news on Social Networks or Messaging
platforms than on more traditional (and more regulated) news media such as
newspapers, radio or TV. Unfortunately, those managing social networks are not
responsible for the contents distributed through their networks, so false or misleading
messages cannot be punished. Only in more extreme cases in which a user violates
the privacy of a person or an institution, that user, but not the network, will be sued
\cite{kekulluoglu2018preserving}.
However, the society has experienced an increasing sensitivity about the contents
spread through social networks. Specially after the concerns about the effects of
fake news circulating social media on the results of the 2016 US presidential
elections \cite{allcott2017social} . Although people are more likely to believe the news that favor their
preferred candidate or show disapprobation for the opponent, it is generally accepted
that fake news have an impact on the way people think, so the impact of social
networks on elections in the US and other countries has been analyzed extensively
\cite{badawy2018analyzing}\cite{borondo2012characterizing} .
Social Media managers have detected a threat of missing customers due to distrust
on social media due to a lack of verification of the contents, that was also amplified
by privacy concerns over politically related data breaches \cite{venturini2019api}. Therefore there is a
growing interest on managing and controlling rumor dissemination on social media,
which starts by detecting fake news on social media \cite{shu2017fake}.

In this paper, we propose a causal detection approach to tackle fake news classification, we will only focus 
in retrieved causal relations of text and in giving a probabilistic quantity for an information to be faked
based in previous learn information of trust worthy sources. The paper is organized as follows: Section 2
will briefly describe causality and related work. Then, Section 3 will dive into the proposal of our method
to classify fake news. We provide empirical evidence to support our hypothesis that fake news can be classified
with respect to their causal relations in Section 4 with a set of synthetic experiments and a real experiment.
Finally, we conclude our article with a conclusions and further work section.

\section{Causality applied to fake news and related work}
Fake news detection is problematic since it appears in any context. Then, in contrast with more concrete
problems such as automatic entity recognition, it is very difficult to extract patterns from texts in a supervised machine learning fashion, tagging doubtful messages, in a similar way as spam detectors label email messages with a calculated spam probability. Nevertheless, the plethora of different contexts for fake news, sharing between them almost any pattern, is too broad for pure machine learning systems \cite{khan2010review} to perform a good work. 

Motivated by the abscence of a proper approach to tackle fake news and by the difficulties that machine learning systems experiment in this task, we have decided to focus on causality to discriminate if a processed text contains fake information. We can only know if an information is fake based in previous learned information that is accurate. The difficulty is that this information is different for every context. Hence, we must learn from every possible context with before classifying if an information is fake. In this paper, we have decided to learn the causal relations of trust worthy texts by creating a casual weighted graph \cite{garridomerchn2020uncertainty} and then using that information to discriminate if a new information is fake.

Causality is an important notion in every field of science. In empirical sciences, causality is a useful way to generate knowledge and provide explanations. When Newton’s apple fell on into his head, he discovered gravity analysing what had provoked such event, that is, he established a cause-effect relationship \cite{hausman1998causal}.

Causation is a kind of relationship between two entities: cause and effect. The cause provokes an effect, and the effect is a consequence of the cause. Causality is a direct process when A causes B and B is a direct effect of A, or an indirect one when A causes C through B,and C is an indirect effect of A \cite{aish2002explanatory}. In this work, we use causality as source to verify information, as causal sentences establish strong relations between concepts easy to validate.

Causality is not only a matter of causal statements, but also of conditional sentences. In conditional statements, causality emerges generally from the relationship between antecedent and consequence. In \cite{puente2010extraction}, Puente et al. described a procedure to automatically display a causal graph from medical knowledge included in several medical texts, particularly those identified by the presence of certain interrogative particles \cite{sobrino2014extracting}. These sentences are pre-processed in a convenient way in order to achieve single cause-effect structures, or causal chains from a prior cause to a final effect. Some medical texts, adapted to the described process, let make questions that provide automatic answers based on causal relations. Causal links admit qualifications in terms of weighting the intensity of the cause or the amount of links connecting causes to effects. A formalism that combines degrees of truth and McCulloch-Pitts cells permits us to weigh the effect with a value; so, to validate information in a certain degree. In the following section, we propose a classifier based in a causal weighted graph model to predict fake news based in a causality methodology.

\section{Generating a Weighted Causal Graph from Text Causal Relations}
In this section, we will expose the mechanism that detects fake news by means of a Weighted Causal Graph \cite{garridomerchn2020uncertainty}. The weighted causal graph is a generalization of a causal graph \cite{merchan2019generating} but modelling the certain degree as a latent variable with a probability distribution instead of with a point estimation. This model builds a causal network from the logicist point of view of probability with probability distributions in the edges that represent the probability of the certain factors that connect two concepts in the net. This weighted causal graph was specifically designed to retain the uncertainty given by time adverbs of retrieved causal sentences from texts. The priors for each adverb belong to the exponential family of distributions. Further information about the model can be consulted in this paper \cite{garridomerchn2020uncertainty}.

In order to compute the distance between two distributions $P$ and $Q$ we use the KL divergence, that is given by the following expression, where we approximate the integral by Grid Approximation:
\begin{align}
D_{KL}(P || Q) = \int_{\infty}^{\infty} p(x)\log(\frac{p(x)}{q(x)}) dx.
\end{align}
These quantity, along with entropy, will be used in the following section to get a probabilistic prediction of a new causal relation representing fake knowledge.
\section{Detecting fake news}
After performing the learning process of the graph, we are more sure about the uncertainty that a given concept is associated with some effect. As the causal relation may appear in more texts, we need a mechanism to ensure that the new texts contain information that can be trusted. By working with PDFs, we can use the entropy of the PDFs as a measure of the gained information. We compute the entropy of the computed posteriors by grid approximation:
\begin{align}
h(X) = - \int_{\mathcal{X}} f(x) \log f(x) dx
\end{align}
In order to detect fake information, we first need to know if we have gained knowledge about the probability of the causal. The causal is represented by its prior distribution $p(x)$ and posterior distribution $s(x)$. A criterion to discriminate if we have gained knowledge about the probability of the causal is to define a binary variable $b$ that tells us whether it makes sense to compute the fake information probability:
\begin{align}
b =  1-\delta(h(s(x)) - h(p(x))).
\end{align}
If the entropy of the posterior is lower than the entropy of the prior, we have gained knowledge about the probability of the causal. Then, we can compute a fake information probability. We assume that if the new PDF is close to the learned posterior, the information can be trusted. We also take into account the gained information of the posterior. We are more sure about the fake new probability of the distribution $l(x)$ if the entropy of the posterior is close to zero, as we have gained more knowledge. 

By combining both criteria, we propose a criterion for a causal to be fake. First, we need to normalize the KL $KL_n(s(x) || l(x))$ and the entropy $h_n(s(x)) \in [0,1]$. The entropy can be normalized with respect to the other adverb PDFs in a $[0,1]$ range in order to output a probability $p_f(x) \in [0,1]$. KL is more challening to be normalized as it is a measure that has no upper limit and is not defined with distributions that have support $0$. In order to circunvent these issues, we smooth the distributions by a very small quantity and transform the KL divergence by squashing it into a sigmoid. That is, $KL(s(x)||l(x)) = 1-\exp(-KL(s(x)||l(x)))$. By performing these two operations, we get rid of computational problems. We also assign a regularizer weight $w \in [0,1]$ to the entropy factor $h_n(s(x))$ in order to model the contribution of the posterior entropy to the probability $p_f(x)$ and a scale factor $\sigma$ to $p_f(x) \in [0,1]$ that scales the probability in order to ensure it to be fake or to be more permissive with the result. The final criterion, $p_f(x)$, that determines whether a causal relation is fake or not is given by:
\begin{align}
p_f(x) = ((1-w)(KL_n(s(x) || l(x)) w h_n(s(x))))^{\sigma}.
\end{align}
The computed probability can be used for two things. First, it is useful to compute this probability and to discriminate if the analyzed information represent fake information. Hence, it can be used as a fake new classifier. We can train our classifier with a reliable source of information with respect to some particular scenario, for example, lung cancer causal relations. Once we have trained our classifier in a realiable source of information, our posteriors represent the learned knowledge.

At this point, we are ready to discriminate whether new information is fake. We define fake information to be the information that is not similar to the learned information. If the causals retrieved from new sources of information give a high fake information probability $p_f(x)$, then we can classify this information as fake. We classify information as fake if the probability $p_f(x)$ is higher than a threshold $\beta$ that we introduce as a hyperparameter. For example, if we set $\beta = 0.6$ and $p_f(x) = 0.7$, the information will be fake. On the other hand, if $p_f(x)=0.5$, the informaiton is not fake. By setting this threshold we do not only give a probability but a decision about the new information. We can also give a confidence degree of our decision $\Omega$ by computing $\Omega = |p_f(x)-\beta|/max(1-p_f(x),p_f(x))$ where $\Omega \in [0,1]$ will also represent a probability.

The described expressions represent whether a given causal is fake and also provide its probability. Another interesting measure to compute is the probability of a fake source of information. This probability represent whether the whole source of information is reliable or not. This probability will be computed as a function of the probabilities of being fake of all the causal relations retrieved from the new source of information. The trust degree of a new source of information $\alpha$ can be computed as: $\alpha = \frac{\sum_{i=1}^{N} p_i}{N}$, where $N$ is the total number of causal relations and $p_i$ represent the probability of a causal relation $i$ for being fake. By computing this probability we can also set another threshold $\gamma$ and compute a confidence degree of our decision $\epsilon$ that will clasify is a source of information is reliable in an analogous form that we have computed it for a single causal relation. That is, $\epsilon = |\alpha -\gamma|/max(1-\alpha,\alpha)$ computes the confidence degree of our decision to classify a source of information.

The second use for the probability of a causal relation as fake is to serve as a binary variable for learning new knowledge. We can refine our classifier with new knowledge as some knowledge is refined as a function of time. The old knowledge may not serve as time passes, so we need to learn knowledge from new sources of information. The problem that arises is to discriminate whether the new knowledge can be trusted. In order to discriminate if the new knowledge is trustworthy we do only have the learned posteriors, the probability of a new causal relation to be fake and the probability of a source of information of being reliable.

\begin{figure}[htb]
\begin{center}
\includegraphics[width=0.7\linewidth]{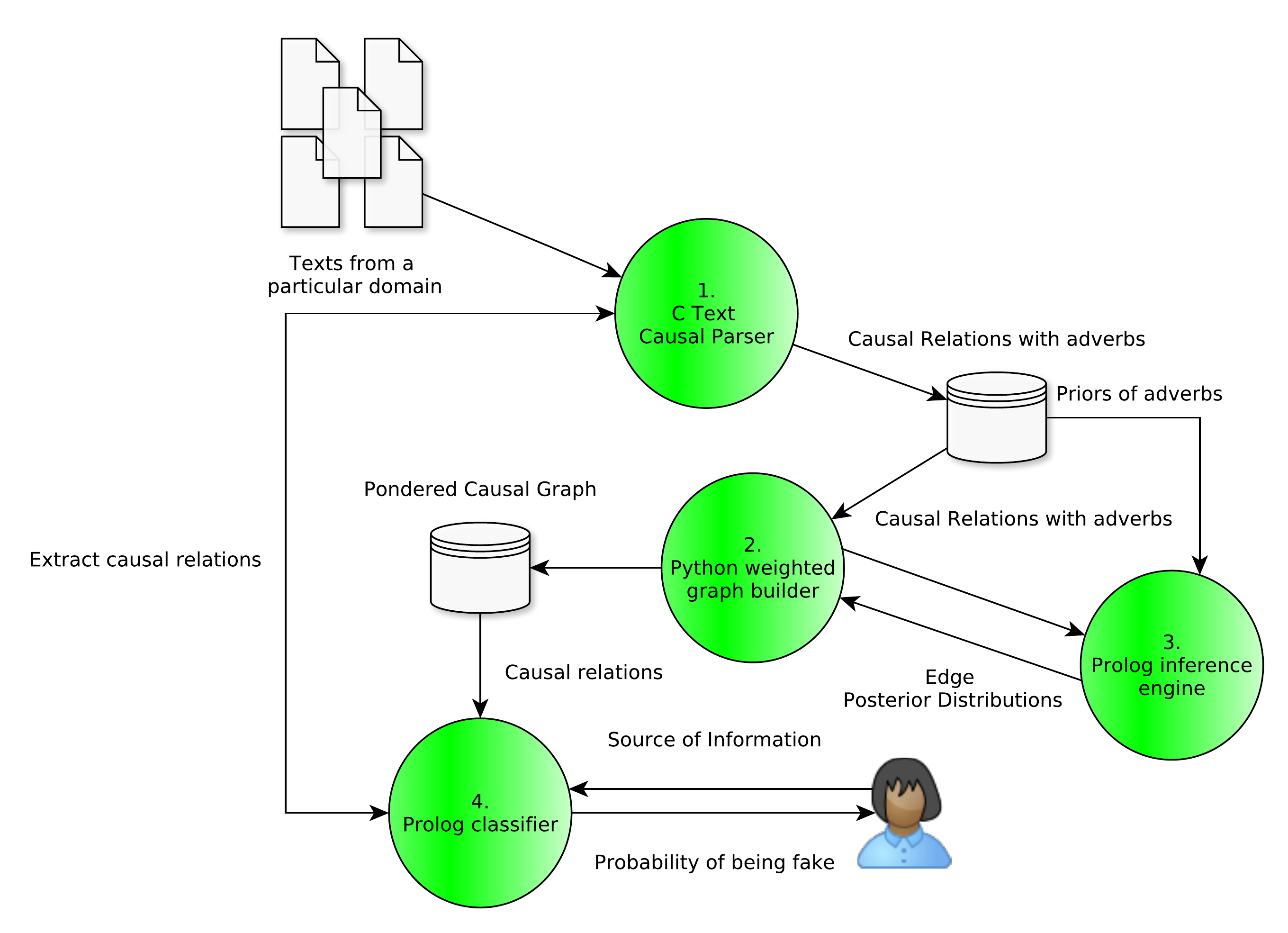}
\caption{Architecture of the fake news classifier. First, causal relations are retrieved with a parser from texts belonging to a particular domain. Then, a probabilistic causal graph is generated with all the causal relations retrieved. By using this probabilistic causal graph, we can build a fake news classifier that reads some input information and outputs the probability of that information to be fake.}
\label{fig:system}
\end{center}
\end{figure}

In this case, we recommend to learn new causal relations if the source of information is trustworthy, ignoring the single probabilities of the causal relations. The probability of a source of information being trust worthy is a function of every single causal relation contained in it, so if the causal relations lie far away from our posteriors, the source of information will be classified as not trust worthy and we will learn nothing. On the other hand, if we learn only the causal relation that are not fake by only focusing on the probability of every single causal relation we will potentially not learn nothing new about the context of the text that we are analyzing. We can adjust the rate of change in the posterior distributions with the mentioned threshold $\gamma$. However, our methodology can also accept to learn only the causal relations which probability $p_f(x)$ is above the threshold $\beta$ that we can configure for classifying if a causal relation is fake or not. We can also be more restrictive in the learning process by only learning a causal relation or a new source of information if our confidence degree $\Omega$ or $\epsilon$ respectively lies above other thresholds that we can configure. We have designed a flow diagram that is represented in Figure \ref{fig:system} that illustrates all the information of our system.
\vspace{-.5cm}
\section{Experiments}
In this section, we will present several experiments where the proposed methodology tackles with new knowledge to be learned. We have performed a set of different synthetic experiments to illustrate the usefulness of the proposed approach. We also present a real case involving Lung Cancer from a trained weighted causal graph from a trusty source. All the code for the experiments and the fake news classifier is available in \url{https://github.com/EduardoGarrido90/prob_causal_graphs_fake_news}.

In all our experiments, we set the following value to the hyperparameters of our model: We assign $w$ value $0.2$ in order to give to the entropy factor a $20\%$ of importance and to the KL divergence factor a $80\%$ of importance for the final expression. We consider that the divergence between distributions is more important than the decrease in entropy since we are more interested in detecting fake information that in being sure about whether our learned information is robust and coherent. Depending on the application, this value can be changed to meet the user preferences. We have also set $\sigma$ to $3$, since we have empirically observed that this value classifies fake information accurately.
\subsection{Synthetic Experiments}
We generate synthetic data that contain causal relations of the form \texttt{A->B} and \texttt{B->C}. Every generated causal relation comes with a time adverb that has a prior probability distribution. The prior distribution for every different causal relation is the prior distribution associated with the adverb that comes with the first generated causal relation. We generate four scenarios that illustrate the different outputs that our system can consider. For every scenario, we first generate a train set that represent the source of truthful information and a test set that represent the new source of information. In each synthetic experiment, we generate a number of causal relations automatically, choosing the adverbs of the causal relations at random from a subset of all the possible time adverbs that our system considers, and retain only those causal relations of the form \texttt{A->B} and \texttt{B->C} from all possible combinations of the set $\{A,B,C\}$. If it is not said otherwise, we consider a threshold of $30$ to consider an information not being fake.

\begin{figure}[htb]
\begin{center}
\begin{tabular}{ccc}
\includegraphics[width=0.33\linewidth]{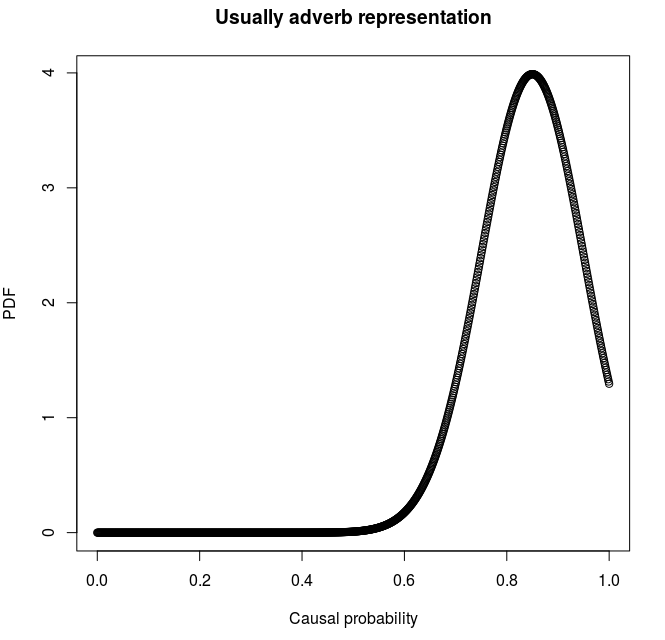}&
\includegraphics[width=0.33\linewidth]{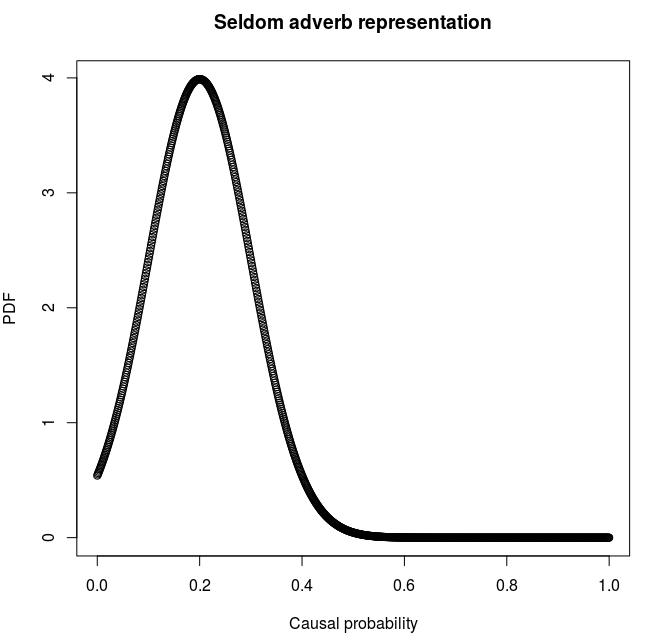}&
\includegraphics[width=0.33\linewidth]{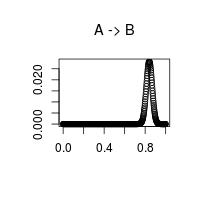}
\end{tabular}
\caption{Prior (left) and posterior (right) distribution of a synthetic causal relation. The new knowledge that we want to validate is represented by the figure of the center. This relation is rejected due to the fact that it is very different from the learned knowledge.}
\label{fig:fse}
\end{center}
\end{figure}

For the first synthetic experiment, we generate $200$ causal relations with the adverbs usually and normally for the train set and retrieve only the valid ones according to the described criterion. For the test set, we generate another $5$ causal relations but this time with the adverbs infrequently and seldom, whose probability distributions are very different from the ones of the train set as we can see in the figures of the left and center of the Figure \ref{fig:fse}. The distributions of the train set generate a posterior distribution that is very different from the posterior distributions of the train set as we see in the right figure of Figure \ref{fig:fse}.
We expect that our system consider the test set of knowledge to be fake information. We execute the Prolog inference engine, obtaining the probabilities in the range of $[78\%,80\%]$ for every causal relation of the test set of being fake causal relations.

We can observe that the system inferred that the new information is fake with a high probability. This satisfies our hypothesis that if we consider probability distributions different from the learned ones, the system would argue that the information would be fake. We would also like to analyze whether the source of information must be learned as a whole or not. As the causal relations are considered to be fake with high probability, the system will have to reject the new source of information. Our proposed system outputs the following log once it analyzes the test set: \textit{The probability of the source being non trust worthy is : 78.557881\%. According to the given threshold, we must not learn causal relations from this source. The confidence degree of the decision based in the threshold and the probability of the source is 55.446863\%.}

As we can see, the system does not recommend to learn the information given by the test set, which validates again our proposed hypothesis.

\begin{figure}[htb]
\begin{center}
\begin{tabular}{ccc}
\includegraphics[width=0.33\linewidth]{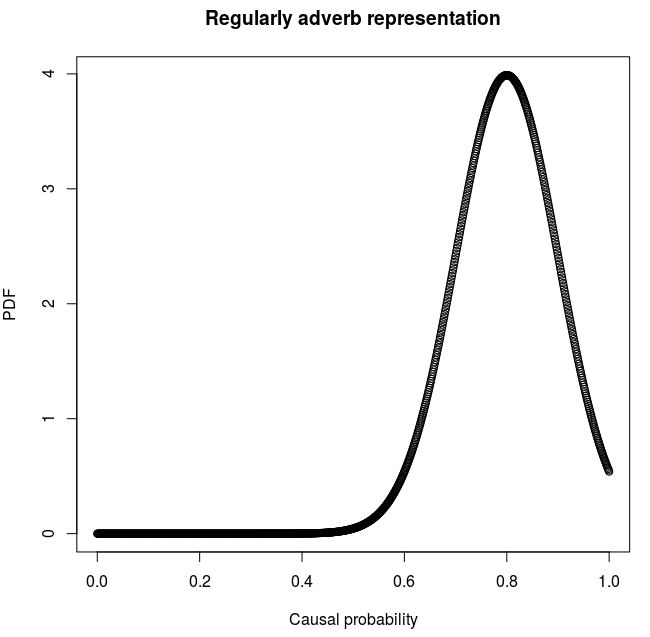}&
\includegraphics[width=0.33\linewidth]{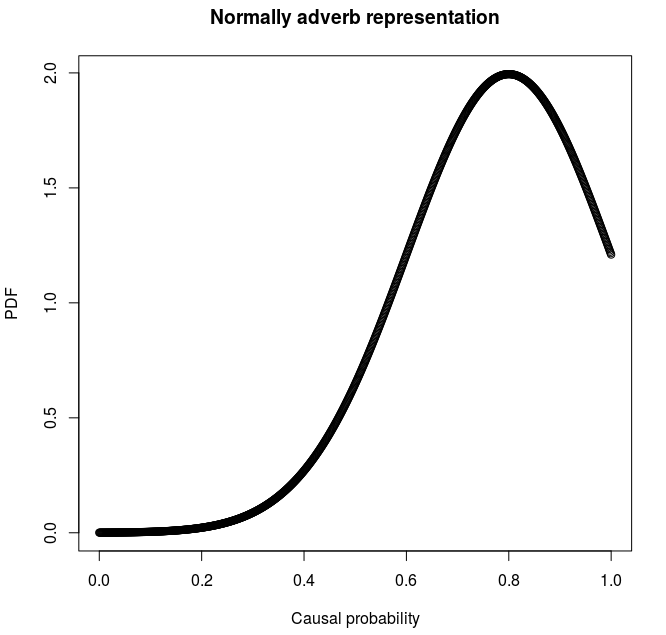}&
\includegraphics[width=0.33\linewidth]{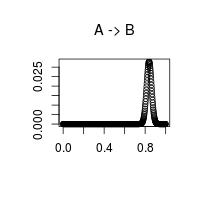}
\end{tabular}
\caption{Prior (left) and posterior (right) distribution of a synthetic causal relation. The new knowledge that we want to validate is represented by the figure of the center. This relation is accepted due to the fact that it is similar to the learned knowledge.}
\label{fig:sse}
\end{center}
\end{figure}

The second experiment is just an example of the opposite scenario, an scenario where the new information must be learned. In order to simulate this case, we generated for the train set causal relations with the adverbs usually and normally and for the test causal relations with the adverbs frequently and regularly as we can see in the figures of the left and center of Figure \ref{fig:sse}. The distributions of the train set generate a posterior distribution that is similar to the distributions of the test set, as we can see in the figure of the right of Figure \ref{fig:sse}.
Based on this data, we expect a low probability for the new causal relations of being fake, being able to be incorporated in the weighted causal graph. The system compute the following probabilities for the $5$ causal relations considered for the test set: $[18.76\%, 26.69\%, 54.12\%, 18.77\%, 46.7\%]$.

We see how the majority of the causal relations lie behind the considered $30 \%$ threshold. Some relations are suspicious of being fake though. In order to know whether we can learn from this test set, we analyze the source of information as a whole, obtaining the following results: \textit{The probability of the source being non trust worthy is : 33.010959\%. According to the given threshold, it is a trust worthy source. The confidence degree of the decision based in the threshold and the probability of the source is 2.969203\%.}

We observe how our system recommends to learn the information from the test set but with a low security grade. The security grade can vary according to the given threshold and to the hyperparameters $\sigma$ and $w$. For these experiments we have configure the system in an strict way to not consider information suspicious of being fake. Higher values for $w$, $\sigma$ and lower values for the threshold relax the system criterion.
\subsection{Real Experiment}
The proposed methodology has been applied to detect fake news in a medical environment. The Mayo Clinic is a prestigious healthcare institution based in Rochester, Minnesota, very well known for its activities in Cancer Research. They developed a database of Tweets related with lung cancer, and labeled those messages as real of fake, in an effort to mitigate the effect of fake messages on patients. 

Cancer patients, especially those recently diagnosed, are very vulnerable to any type of fake messages \cite{oh2019people}. The anxiety of the patients (and their families) to know why they have become sick or the possible methods to increase the chances of a quick recovery, make some patients immerse themselves in Social Networks and web pages searching for answers. Fake messages about origins of the disease, naturalistic magic cures, unproved diet methods, etc. may have a very negative impact of patients including an increasing anxiety or lack of trust in standard medical treatments.

We use our system in the Lung Cancer domain trained from the Mayo clinic text in order to classify tweets as fake information or not. Twitter is a source of information where the society can talk freely about any topic exposing opinions that are not contrasted from information sources validated by experts \cite{weller2014twitter}. Hence, classifying information from Twitter based in information from the Mayo clinic is an excellent example where our system can operate. 

\begin{figure}[htb]
\begin{center}
\includegraphics[width=0.99\linewidth]{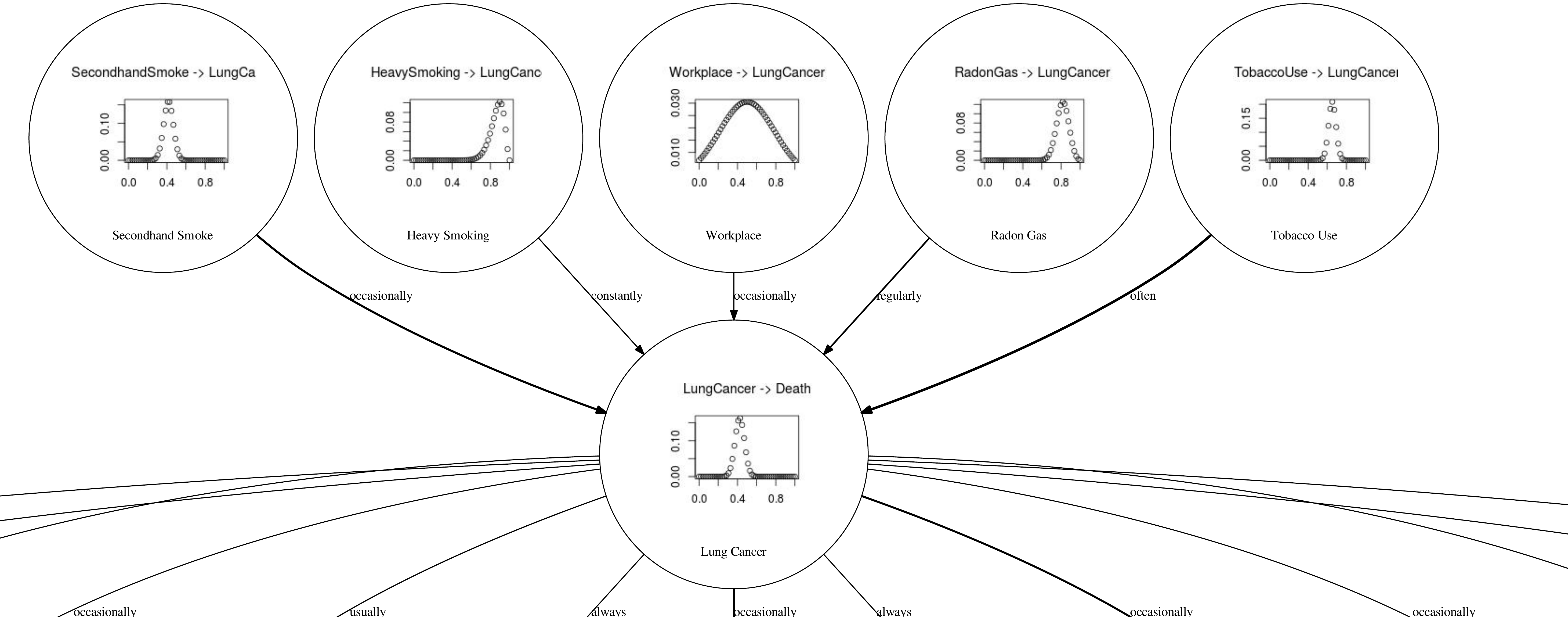}
\caption{Posterior distributions of the causes of Lung Cancer of the causal weighted graph in the Lung Cancer domain.}
\label{fig:tre}
\end{center}
\end{figure}

We have learned the following causes of lung cancer from the Mayo Clinic set of documents that are shown in Figure \ref{fig:tre}.
Searching what do the users write about lung cancer and its causes, we retrieve the following two tweets that contain causal information: 1. \textit{But this wasn't always the case. Today, smoking causes nearly 9 out of 10 lung cancer deaths, while radon gas, pollution, and other things play a smaller role} and 2. \textit{Cigs = lung cancer \& DEATH}. These two tweets contain the following causal relations: 1. \textit{Smoking constantly causes lung cancer}, 2. \textit{Radon gas hardly ever causes lung cancer}, 3. \textit{Secondhand smoke hardly ever causes lung cancer}, 4. \textit{Smoking always causes lung cancer} and 5. \textit{Lung cancer always causes death}. 

\begin{figure}[htb]
\begin{center}
\begin{tabular}{cc}
\includegraphics[width=0.45\linewidth]{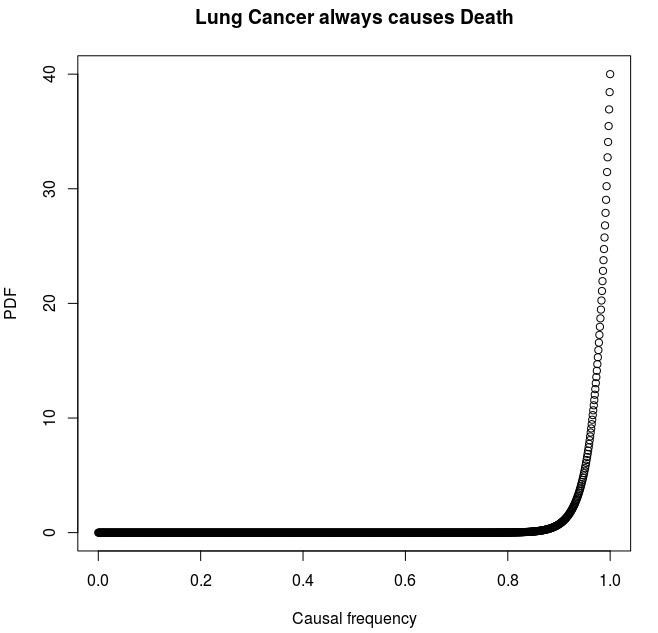}&
\includegraphics[width=0.45\linewidth]{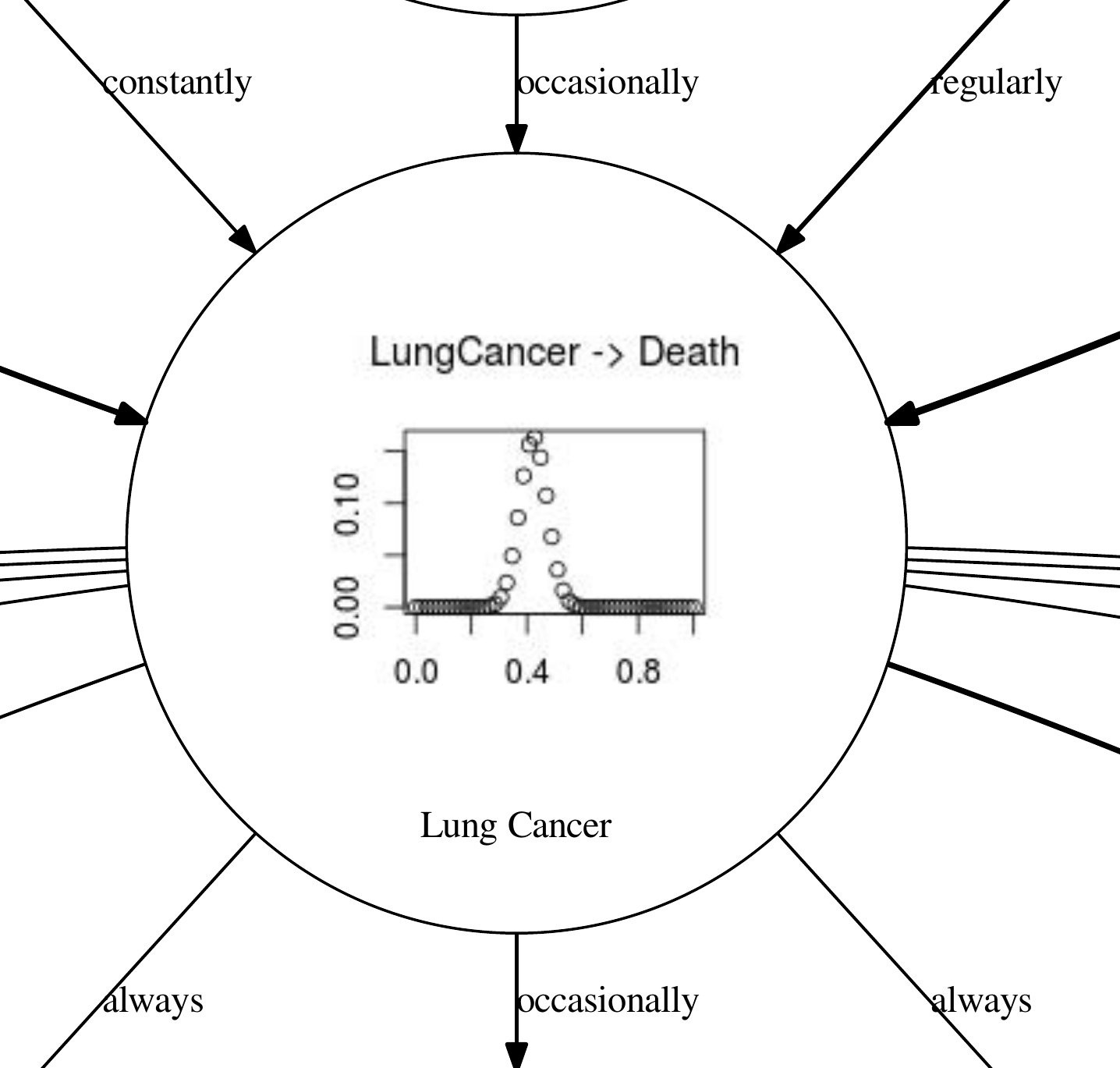}
\end{tabular}
\caption{Comparison of a causal relation to be learned (left) and the posterior distribution of a causal relation of the weighted graph (right).}
\label{fig:sre}
\end{center}
\end{figure}

We are going to execute our algorithm to discriminate whether the new information is fake or not. We can hypothesize that the information is going to be fake as it can be argued that is slightly radical, which is common in social networks. We can observe this radicality by comparing the posterior distribution of lung cancer and death of our system and the distribution associated with the analyzed information of Twitter in Figure \ref{fig:sre}.
We can see in Figure \ref{fig:sre} that the distribution to be learned is very different to the posterior distribution of the weighted graph, so we can not learn it. Fake probabilities are: $[49.39\%, 65.66\%, 55.94\%, 71.46\%, 68.58\%]$.

We can observe that all the probabilities are high, hence, the system must classify the source of information as fake. If we do classify this source of information, the system generates the following log: \textit{The probability of the source being non trust worthy is : 62.203338\%. According to the given threshold, we must not learn causal relations from this source. The confidence degree of the decision based in the threshold and the probability of the source is 35.694769\%.}

And refuses to learn the information found in Twitter, as we have hypothesize that it would do. This example has shown how we can use the described system to determine whether information is fake or not.
\section{Conclusions and Further Work}
We have presented a system that classifies if new knowledge must be learned based in if it represents fake information, which is a problem in our society. Our system bases its decision in a pondered causal graph that has been generated from causal relations retrieved from texts. The system can discriminate whether new sources of information are worth to be added in the pondered causal graph. We obtain as an output a probability that represents if the new information represents a fake new and a security grade of the decision. We consider as further work to optimize the three hyperparameters, the threshold, $w$ and $\sigma$; of the proposed model with Multiobjective Constrained Bayesian Optimization \cite{garrido2019predictive}. More further work involves to feed in a cognitive architecture of a robot the model, so its information can be shared with other robots to see how they update its networks. The robots would question and answer queries, updating their networks and showing social behaviour that, combined with other processes, can be a correlation of machine consciousness \cite{merchan2020machine}. We also want to compare this approach with a machine learning generic NLP model such as BERT \cite{devlin2018bert}. We hypothesize a better performance of our approach, since it is able to structure causal information coming from different texts.
\section*{Acknowledgments}
The authors gratefully acknowledge the use of the facilities of Centro
de Computaci\'on Cient\'ifica (CCC) at Universidad Aut\'onoma de
Madrid. The authors also acknowledge financial support from Spanish
Plan Nacional I+D+i, grants TIN2016-76406-P and TEC2016-81900-REDT.
\bibliographystyle{acm}
\bibliography{notes}
\end{document}